\documentclass[11pt]{article}

\usepackage[final]{acl}

\usepackage{times}
\usepackage{latexsym}
\usepackage{booktabs}
\usepackage{hyperref}
\usepackage[T1]{fontenc}

\usepackage[utf8]{inputenc}

\usepackage{microtype}

\usepackage{inconsolata}

\usepackage{graphicx}

\usepackage{makecell}
\usepackage{multirow}
\usepackage{color, colortbl}
\usepackage{amsmath}
\definecolor{Gray}{gray}{0.9}

\title{Measuring Artifact Reliance in Machine-Generated Text Detectors}

\author{Anastasia Voznyuk \\
  Advacheck OÜ \\ Tallinn, Estonia \\
  \texttt{voznyuk@advacheck.com} \\\And
  Andrey Grabovoy \\
  Moscow State University \\   Moscow, Russia \\
  \texttt{andriy.graboviy@gmail.com} \\
}

\begin{document}
\maketitle
\begin{abstract}
Detectors of machine-generated text score almost perfectly on scientific-literature benchmarks, implying we can reliably distinguish with these detectors human and machine-generated documents. We looked at ten scientific-document subsets from six popular resources, and on several of them that apparent ease is a side effect of how the documents were put together. A logistic regression using $27$ superficial features reached $0.833$-$0.997$ macro-$F_1$, and on the three arXiv subsets it outperformed every detector we evaluate with $F_1$-scores $0.951$–$0.997$. The strongest cue to the detector is the PDF extraction artifact: human texts retain their original line breaks, whereas generated texts are extracted as a single block of prose. This cue, together with three other formatting features, tell the classes apart at $0.996$ AUC. To distinguish benchmarks that contain such a signal from detectors that rely on one, we introduce format counterfactuals: just re-wrapping a machine-written abstract without changing a single word reverses $20$-$27\%$ of a public detector's correct decisions. Normalizing formatting before training reduces the generalization gap from $0.32$ to $0.22$. On format-clean benchmarks, it also raises out-of-domain macro-$F_1$ from $0.33$ to $0.65$ for detectors trained on artifact-containing subsets. We release our tools and scripts to reconstruct the corrected subsets\footnote{\href{https://github.com/natriistorm/ArtifactsInMGTBenchmarks}{Code to the paper}}.

\end{abstract}

\section{Introduction}

Establishing whether a document was written by a person or a language model is a provenance
problem, and in the scientific literature it has become a practical one: an analysis of 15
million PubMed abstracts estimates that 13.5\% of the papers written in 2024 were LLM-assisted~\cite{doi:10.1126/sciadv.adt3813}. Publishers, reviewers, and archives
consequently need detectors they can trust, and trust in a detector usually depends on the benchmark that measured its performance.

The benchmarks for machine-generated text detection report that the problem is close to be solved. During shared tasks on machine-generated text detection, participants build systems above $0.97$ accuracy (Appendix~\ref{appendix:evaluation}), and the authors of GPABench2 report over $99.9\%$
accuracy on fully GPT-written abstracts~\cite{liu2024detectability}. This is difficult to
reconcile with the fact that generators have improved continuously over the same period: if
the underlying task were being measured, it should have become harder, not easier.

We argue that for several of these benchmarks, a large part of the reported performance is dependent on how the benchmarks were assembled, which often originates in document processing. Human documents in these collections are usually scraped or converted from source formats and retain the formatting of those sources; meanwhile machine documents are emitted directly by an API as clean text blocks. The difference in formatting is often not immediately visible, while being trivially learnable and exploitable. In the arXiv subsets we examine, a classifier over four content-free formatting features
separates the classes at $0.996$ AUC.

Our contributions are as follows:
\begin{enumerate}
    \item We conduct a surface-attribution analysis of ten scientific-document subsets from six resources, showing that content-independent stylistic features can match or exceed the corresponding shared-task winners. We also identify the surface dimensions that carry these shortcuts.
    \item We introduce format counterfactuals and the Artifact Reliance Score: a content-preserving intervention and a model-agnostic measure of a detector's reliance on formatting when making machine-class decisions. It requires neither retraining nor access to model internals.
    \item We show that artifact exploitability is tokenizer-dependent: SentencePiece-based models discard the whitespace cue, so detector performance scores on these benchmarks are not comparable across tokenizer families.
    \item We show that normalizing formatting before training improves cross-benchmark generalization by reducing reliance on document-construction artifacts, including when detectors trained on artifact-containing datasets are evaluated on format-clean benchmarks.

\end{enumerate}

\section{Related Work}
\label{sec:task_methods}

\paragraph{Detection methods.} Machine-generated text detection is usually formulated as binary
classification of human and machine authorship~\cite{Fraser2025}. There are three families of methods that are usually used for this task: first, statistical and stylometric methods that use token-probability or
linguistic features~\cite{gehrmann-etal-2019-gltr, jawahar-etal-2020-automatic}.
Then, zero-shot probability methods threshold text-level scores such as
perplexity or log-rank~\cite{su-etal-2023-detectllm, hans2024spotting}, or compare
statistics of perturbed and original text~\cite{mitchell2023detectgpt}. Finally, fine-tuned language models (e.g.\ RoBERTa, DeBERTa) are the current state of the art~\cite{macko-etal-2023-multitude, he2023debertav} but dependent on substantial labelled data.

\paragraph{Artifacts and benchmark difficulty.} Several surveys catalogue MGT detection
benchmarks~\cite{Fraser2025, wu2023survey}, but only few examine the \emph{quality} of
the data itself, despite its direct effect on detector reliability~\cite{gritsai2025aidetectorsgoodenough}. That detection benchmarks may be separable
for the wrong reasons is not a new suspicion. \citet{dugan-etal-2024-raid} report formatting
confounds in RAID, naming numbered lists in generated recipes and periodic newlines as such artifacts and estimate their effect on detection as small;  on the scientific subsets we
examine, the same class of cue is instead dominant, reaching $0.996$ AUC and exceeding a fine-tuned transformer. \citet{ayoobi2025easywinstexthardnessaware} 
argue then more broadly that existing benchmarks over-represent easy examples and propose
hardness-aware construction. Our framing follows the broader dataset-artifact literature in NLU, where annotation artifacts allow hypothesis-only models to
succeed~\cite{gururangan-etal-2018-annotation} and motivate bias-only debiasing
methods~\cite{clark-etal-2019-dont, karimi-mahabadi-etal-2020-end}.

\paragraph{Shared tasks.} MGT shared tasks have driven progress by releasing
curated data and evaluation protocols across multilingual, cross-domain, and
cross-generator settings~\cite{dagpap22, pan2024, semeval2024task8}. Some participants of the shared tasks apart from detectors also contribute secondary analyzes of given datasets focused on identifying construction biases, false-positive rates on human texts, and repetitive patterns in synthetic outputs~\cite{voznyuk-konovalov-2024-deeppavlov}, which motivate the
dataset-quality focus of this work.

\section{Benchmarks Analyzed}
\label{section:data}

We analyze ten subsets of datasets with scientific documents drawn from six published resources. Datasets were selected on four criteria: the documents are
scientific artifacts (paper abstracts, paper introductions, or peer reviews); the task is
binary human-vs-machine classification; both classes are present in the release; and the
machine class is written from scratch from a title or prompt. The last criterion matters because rewriting a human text conditions the generation on human writing and yields a
distribution closer to human and AI co-writing than to independent generation
\citep{mixset}. Where a resource offers several generation regimes we therefore keep only the from-scratch one: for example, for GPABench2 we use the \textsc{gpt-write}
task, and for CHEAT we use the
\textit{generation} split. All selected subsets are in English. See Appendix~\ref{appendix:data} for more details on each chosen dataset.

\paragraph{The M4 family.} Six of the ten subsets descend from a single family of shared-task resources.
First, from \textbf{M4}~\citep{wang-etal-2024-m4} we take texts on two scientific domains, arXiv abstracts and peer reviews adapted from PeerRead, \citealp{kang-etal-2018-dataset}. \textbf{SemEval-2024
Task 8}~\citep{semeval2024task8} built its detection-related subtasks on M4, and \textbf{COLING-2025 GenAI Task
1}~\citep{wang2025genai} in turn extended that data with more recent generators, including GPT-4o, LLaMA-3, Gemma-2, and Mixtral.. Each generation therefore inherits its human counterpart from the previous release.

\paragraph{Full scientific papers.}
Another analyzed dataset, \textbf{IDMGSP}~\citet{mosca-etal-2023-distinguishing}, collectes full machine-generated
papers and not just abstracts, pairing human articles from arXiv with
generations from SCIgen, GPT-2, GPT-3, ChatGPT, and Galactica, and splitting each record
into title, abstract, introduction, and conclusion fields. We use the introduction, as it is usually the longest yet representative of a document.

\paragraph{Large-scale abstract collections.}
\textbf{GPABench2}~\citep{liu2024detectability} contains human, GPT-written,
GPT-completed, and GPT-polished abstracts across computer science, physics, and the
humanities and social sciences, generated with four prompt templates per task. We kept only computer science and physics subsets.
\textbf{CHEAT}~\citep{yu2025cheat} provides title--abstract pairs for computer-science
papers with human abstracts collected from IEEE Xplore, and ChatGPT abstracts written from the title and keywords.

\section{Diagnosis}
\label{sec:diagnosis}

\begin{table*}[t]
\centering\small
\setlength{\tabcolsep}{6pt} 
\caption{Surface probe. Raw and Norm. are
macro-$F_1$ before and after format normalization.
\textsc{fmt-only} and \textsc{len-only} are AUCs of the four formatting features, and of
document length alone. TF-IDF is a lexical reference on the same folds. The three arXiv
subsets lose $0.20$-$0.27$ $F_1$ to normalization; the other lose nothing.}
\label{tab:probe}
\begin{tabular}{lrcccccc}
\toprule
Subset & $n$ & Raw & Norm. & $\Delta$ &
\textsc{fmt-only} & \textsc{len-only} & TF-IDF \\
\midrule
SemEval24 arXiv    & 20{,}000 & \textbf{0.997} & 0.729 & \textbf{0.268} & \textbf{0.997} & 0.503 & 0.880 \\
M4 arXiv           & 20{,}000 & \textbf{0.995} & 0.746 & \textbf{0.249} & \textbf{0.996} & 0.635 & 0.917 \\
COLING25 arXiv     & 20{,}000 & 0.951 & 0.752 & \textbf{0.199} & 0.943 & 0.613 & 0.916 \\
\midrule
M4 PeerRead        &  5{,}860 & 0.959 & 0.954 & 0.005 & 0.822 & 0.780 & 0.998 \\
SemEval24 PeerRead &  4{,}714 & 0.941 & 0.914 & 0.027 & 0.709 & 0.541 & 0.947 \\
COLING25 PeerRead  &  6{,}600 & 0.854 & 0.839 & 0.015 & 0.708 & 0.488 & 0.967 \\
IDMGSP             & 16{,}000 & 0.907 & 0.888 & 0.019 & 0.785 & 0.777 & 0.961 \\
GPABench2 CS       & 20{,}000 & 0.940 & 0.938 & 0.002 & 0.547 & 0.596 & \textbf{0.999} \\
GPABench2 Physics  & 20{,}000 & 0.935 & 0.934 & 0.001 & 0.524 & 0.560 & \textbf{0.999} \\
CHEAT              & 20{,}000 & 0.833 & 0.831 & 0.002 & 0.519 & 0.669 & 0.992 \\
\bottomrule
\end{tabular}
\end{table*}

We first establish that the benchmarks contain exploitable patterns, and identify what they
are. We used balanced samples of up to $10{,}000$ documents per class
per subset, with $5$-fold cross-validation and $1000$-sample bootstrap intervals.

\subsection{Surface Probe}
\label{method:probe}

We fit a logistic regression with $27$ content-free surface features: document, sentence and word
lengths, type--token ratio, and per-character rates of punctuation, digits, uppercase,
whitespace runs, newlines, and non-ASCII characters. All features are described in Appendix~\ref{appendix:hyperparameters}. No feature uses actual word content, so the
classifier cannot exploit topic or vocabulary. We report its macro-$F_1$ as an estimate of how
far a detector can get on a benchmark without reading the text, and the univariate AUC of each feature as an attribution. The joint score can be used to determine whether a shortcut exists, while the univariate scores points out which surface dimension carries it.

Table~\ref{tab:probe} reports the probe. It reaches
macro-$F_1$ between $0.833$ and $0.997$ without access to a single word identity. For
comparison, the winning system of SemEval-2024 Task 8 (monolingual) reached $0.975$ accuracy and
of COLING-2025 Task 1 (monolingual) $0.831$ macro-$F_1$ (Appendix~\ref{appendix:evaluation}). 

The \textsc{fmt-only} column isolates the four formatting features, namely newline rate,
whitespace-run rate, curly-quote rate and em-dash rate. On the three arXiv subsets these four alone
reach AUC $0.943$-$0.997$, and normalizing the text, which sets these features to chance by
construction, costs the full probe $0.199$-$0.268$ macro-$F_1$. Lexical TF-IDF score after this transformation stays unchanged. On the remaining seven subsets, the same normalization changes almost nothing ($\Delta \le 0.027$).

Therefore, we can observe two distinct artifact families. The
arXiv subsets carry a pipeline artifact, precisely a formatting difference introduced by how the documents were collected. The abstract
collections, GPABench2 and CHEAT, are format-clean (\textsc{fmt-only} AUC $0.52$-$0.55$) yet
almost perfectly separable lexically (TF-IDF $\ge 0.99$), driven by a mean-word-length gap that
no preprocessing can remove. This is an actual stylistic difference between human and machine scientific prose. The distinction defines what detector may learn: pipeline artifacts are spurious by construction, while stylistic
differences give legitimate signal, although even genuine stylometric features can invert across settings and mislead detectors~\cite{gao-etal-2026-personalization}. We also note that document length, often the first suspected confound, is not found anywhere except IDMGSP and M4 peer reviews.

\subsection{Nature of this Artifact}
\label{diag:whitespace}

The formatting artifact has a mechanical explanation. Human arXiv abstracts were collected with
the line wrapping of their source intact, because the median line length in these texts is $76$ characters implying the classic
$80$-column limits. At the same time, generations were returned by an API as a single unwrapped block,
with median line length $479$-$481$. Measured over the full subsets, every human abstract in M4 and
SemEval24 contains a double space, against $2.2\%$ and $2.0\%$ of machine abstracts; newlines appear in $100\%$ of human documents against $53.1\%$ and $66.3\%$ of machine ones. The
predicate of double space \texttt{"\ \ "} in text is thus a $99\%$-accurate classifier for SemEval-2024's arXiv
subtask.

The artifact was partly fixed: in COLING-2025 there is no gap with double-space ($4.3\%$
human against $6.8\%$ machine), but the newline gap remained ($90.6\%$ against $26.3\%$). Then, another example is IDMGSP, where machine introductions carry newlines ($24\%$) and human ones do not ($0\%$). That reversal will be used as a control in Section~\ref{section:results}.

\section{Method}
\label{sec:method}

Section~\ref{sec:diagnosis} determines exploitable
patterns on these benchmarks. It does not establish that any detector uses them: the probe of Section~\ref{method:probe} gives evidence about
the data and not about detectors. Answering the second question requires intervening on the
input, which is what this section develops.

\subsection{Format Counterfactuals}
\label{method:counterfactual}

\paragraph{Admissible operators.}
We call a text transformation $g$ \emph{admissible} if $g$ alters the formatting / punctuation of the text and never the words
themselves. Admissibility is machine-checkable: source and transformed document must be character-identical once all non-alphanumeric characters are removed, and we verify this on every transformed document we produce. For a formatting dimension $F$ we require an admissible pair: a $\mathrm{strip}_F$ that removes
$F$, and an $\mathrm{impose}_F(\cdot\,;\theta)$ that re-applies it according to a profile
$\theta$ measured from data rather than chosen by us. In this work, we chose $F$ to be the following: let function $\mathrm{strip}$ collapse whitespace runs and newlines to single spaces and folds curly quotes and en/em dashes to ASCII;
then $\mathrm{impose}(\cdot\,;\theta)$ re-wraps lines at line lengths drawn from $\theta$'s empirical
distribution, inserts double spaces after sentence boundaries, and restores curly quotes and
dashes, each at $\theta$'s measured rate. We fit $\theta_H$ and $\theta_M$ separately on the
human and machine documents of each subset, so the counterfactual is calibrated to the benchmarks.

\paragraph{Artifact Reliance Score.}
Let $\mathcal{Y} = \{H, M\}$ and $y(x) \in \mathcal{Y}$ be the gold label of document $x$, $D(x) \in \mathcal{Y}$ a detector's decision, and $s(x) \in [0,1]$ its machine-class score, so that $D(x) = M$ iff $s(x) \ge \tau$. Let $\mathcal{M} = \{x : y(x) = M\}$ be the machine
documents of a subset and $T_{M}=\{x \in \mathcal{M} : D(x) = M\}$ those the detector originally classifies correctly. We then measure how often a formatting-only edit reverses a correct decision, and subtract a placebo edit that imposes
the machine profile instead:
\begin{equation*}
    \mathrm{ARS} = \frac{1}{|T_M|}\sum_{x \in T_M}
        \mathbf{1}\!\left[D(\mathrm{impose}(x; \theta_H)) = H\right]
\end{equation*}
  \begin{equation*}
    \mathrm{ARS}^{\text{plc}} = \frac{1}{|T_M|}\sum_{x \in T_M}
        \mathbf{1}\!\left[D(\mathrm{impose}(x; \theta_M)) = H\right]
\end{equation*}
\begin{equation*}
    \mathrm{ARS}_{\text{net}} = \mathrm{ARS} - \mathrm{ARS}^{\text{plc}}
\end{equation*}

The placebo is an edit of the same kind that does not move the document towards the human class; it allows us to separate artifact reliance from generic instability under formatting edits. Both terms are evaluated on the same documents, so we test the difference with a paired bootstrap over $T_M$.

$\mathrm{ARS}_{\text{net}}$ is the fraction of correct decisions that a formatting edit reverses beyond the placebo, however $T_M$ is not a random sample: for a detector with low machine-class recall it plausibly overrepresents documents whose formatting already resembles the machine profile. Injected human profile moves these documents furthest, so conditioning can inflate the effect it measures. We introduce our primary statistic, the mean drop in the machine score under the counterfactual minus the same drop under the placebo:
\begin{equation*}
    \Delta_{\text{cf}} = \frac{1}{|\mathcal{M}|}\sum_{x \in \mathcal{M}}
      \bigl[s(x) - s(\mathrm{impose}(x; \theta_H))\bigr]
\end{equation*}

\begin{equation*}
    \Delta_{\text{plc}} = \frac{1}{|\mathcal{M}|}\sum_{x \in \mathcal{M}}
      \bigl[s(x) - s(\mathrm{impose}(x; \theta_M))\bigr]
\end{equation*}

\begin{equation*}
    \Delta_{\text{net}} =  \Delta_{\text{cf}} -  \Delta_{\text{plc}}
\end{equation*}
This metric requires no decision threshold, so it is comparable across
the detector families. Also, since it is computed over all of $\mathcal{M}$, it is not inflated by selection. Nevertheless, a detector with low machine-class recall has little probability mass to move, so we report machine-class recall alongside both statistics.

\paragraph{Scope of the operator family.} The construction is not specific to whitespace. Any formatting dimension admits the same
treatment, provided the three conditions above hold: a machine-checkable invariant, a profile
estimable per class, and both transformation directions. Another example of such operator is paragraph and list segmentation, and it is well motivated, since \citet{dugan-etal-2024-raid} report numbered lists
as a comparable structural artifact in generated text. We instantiate whitespace because that is
the dimension the probe selects on these subsets; on the abstract collections it instead selects
mean word length, however, it is a lexical dimension, for which no admissible operator exists.


\section{Experimental Setup}
\label{section:experiments}

We take one representative of each family described in
Section~\ref{sec:task_methods}. For fine-tuned language models we train
\texttt{DeBERTa-v3-base}~\cite{he2023debertav} on each subset. For zero-shot likelihood methods
we use Binoculars~\cite{hans2024spotting}, which compares the perplexity and cross-perplexity of
two models, and Fast-DetectGPT~\cite{fast-detectgpt}, a more efficient variant of
DetectGPT~\cite{mitchell2023detectgpt} based on conditional probability curvature. Neither
zero-shot method requires fine-tuning, which matters in practice because training a detector for
every domain and generator is infeasible.

Per subset we hold out $1000$ balanced test documents and train on up to
$20{,}000$ balanced documents, subject to availability. Hyperparameters and the full grid are given in
Appendix~\ref{appendix:hyperparameters}. For the surface probe and the counterfactual
experiments we use the samples and operators described in Sections~\ref{method:probe}
and~\ref{method:counterfactual}.

\section{Results}
\label{section:results}

\begin{table}[ht!]
    \centering \small
   \caption{Baseline detectors on the ten scientific subsets, as macro-$F_1$ under a
random split: a
fine-tuned DeBERTa-v3-base, Binoculars (Perpl.) and Fast-DetectGPT (Perturb).}
\label{tab:results}
\begin{tabular}{lccc}
\toprule
Dataset & DeBERTa & Perpl. & Perturb. \\
\midrule
CHEAT                & \textbf{0.892} & 0.901 & 0.908 \\
COLING25 (arXiv)     & 0.824 & 0.733 & 0.868 \\
COLING25 (PeerRead)  & 0.680 & 0.804 & 0.946 \\
GPABench2 (CS)       & 0.781 & 0.834 & 0.984 \\
GPABench2 (Physics)  & 0.767 & 0.795 & \textbf{0.998} \\
IDMGSP               & 0.759 & 0.745 & 0.965 \\
M4 (arXiv)           & 0.795 & 0.738 & 0.824 \\
M4 (PeerRead)        & 0.794 & \textbf{0.924} & 0.987 \\
SemEval24 (arXiv)    & 0.769 & 0.702 & 0.786 \\
SemEval24 (PeerRead) & 0.808 & 0.922 & 0.965 \\
\bottomrule
    \end{tabular}
\end{table}

\subsection{Baseline Detectors}
\label{res:detectors}

Table~\ref{tab:results} reports the three baseline detectors on the ten subsets. Best
results are marked per column, but no single column should be read as a verdict on a
dataset; the columns jointly characterise surface separability, behaviour under
perturbation, and difficulty for current detectors. The salient comparison is with the
surface probe of Section~\ref{method:probe}: on the arXiv subsets, against a probe
macro-$F_1$ of $0.951$-$0.997$, the best detector reaches $0.868$. 

\subsection{Artifact Reliance}
\label{res:ars}

\begin{table*}[t]
\centering
\small
\setlength{\tabcolsep}{6pt} 
\caption{Artifact reliance of three public detectors under format counterfactuals.
\textsc{flip} is $\mathrm{ARS}_{\text{net}}$; \textsc{unc} is $\Delta_{\text{net}}$, the placebo-corrected margin shift over all of $\mathcal{M}$; \textsc{rec} is machine-class recall on the original documents ($|T_M|/500$). \textbf{fmt} is formatting separability (Table~\ref{tab:probe}),
by which subsets are ordered.}
\label{tab:ars}
\begin{tabular}{lc|ccr|ccr|ccr}
\toprule
& & \multicolumn{3}{c|}{HC3} & \multicolumn{3}{c|}{GPT-2} & \multicolumn{3}{c}{RADAR} \\
\cmidrule(lr){3-5}\cmidrule(lr){6-8}\cmidrule(lr){9-11}
Subset & fmt & \textsc{rec} & \textsc{flip} & \textsc{unc}
& \textsc{rec} & \textsc{flip} & \textsc{unc}
& \textsc{rec} & \textsc{flip} & \textsc{unc} \\
\midrule
SemEval24 arXiv & 0.997 & 0.318 & 0.453 & \textbf{+0.200} & 0.420 & 0.343 & \textbf{+0.137} & 0.226 & 0.177 & $-$0.006 \\
M4 arXiv        & 0.996 & 0.366 & 0.579 & \textbf{+0.267} & 0.512 & 0.230 & \textbf{+0.121} & 0.334 & 0.054 & $-$0.019 \\
COLING25 arXiv  & 0.943 & 0.444 & 0.423 & \textbf{+0.207} & 0.380 & 0.347 & \textbf{+0.138} & 0.262 & 0.115 & +0.013 \\
M4 PeerRead        & 0.822 & 0.224 & 0.312 & +0.097 & 0.752 & 0.003 & +0.002 & 0.978 & 0.000 & $-$0.002 \\
IDMGSP \emph{(rev.)} & 0.785 & 0.478 & $-$0.339 & $\mathbf{-0.171}$ & 0.862 & $-$0.005 & $-$0.005 & 0.634 & 0.035 & +0.046 \\
SemEval24 PeerRead & 0.709 & 0.456 & 0.171 & +0.109 & 0.628 & 0.051 & +0.023 & 0.794 & 0.013 & +0.022 \\
COLING25 PeerRead  & 0.708 & 0.416 & 0.062 & +0.034 & 0.534 & 0.045 & +0.025 & 0.856 & 0.023 & +0.026 \\
GPABench2 CS   & 0.547 & 0.284 & 0.127 & +0.042 & 0.198 & 0.000 & +0.003 & 0.626 & 0.000 & $-$0.015 \\
GPABench2 Physics & 0.524 & 0.380 & 0.079 & +0.031 & 0.254 & 0.016 & +0.002 & 0.644 & 0.000 & $-$0.008 \\
CHEAT       & 0.519 & 0.632 & $-$0.016 & $-$0.010 & 0.542 & 0.004 & +0.000 & 0.914 & 0.000 & +0.001 \\
\midrule
\textit{arXiv mean} & & 0.376 & 0.485 & \textbf{+0.225} & 0.437 & 0.307 & \textbf{+0.132} & 0.274 & 0.115 & $-$0.004 \\
$\rho$ with fmt & & & +0.770 & \textbf{+0.745} & & +0.552 & +0.600 & & +0.813 & $-$0.103 \\
\bottomrule
\end{tabular}
\end{table*}

We apply the counterfactual of Section~\ref{method:counterfactual} to three publicly available and commonly used
detectors, none trained on these benchmarks: \texttt{chatgpt-detector-roberta}, the detector
released with HC3 dataset~\cite{Su2023HC3PA}; \texttt{roberta-base-openai-detector}, OpenAI's GPT-2 output detector\footnote{\href{https://huggingface.co/openai-community/roberta-base-openai-detector}{openai-community/roberta-base-openai-detector}}; and
RADAR~\citep{RADAR}, which is adversarially trained against a paraphraser. Each is
scored on all $40{,}000$ counterfactual documents, whose admissibility is asserted at construction.

Table~\ref{tab:ars} reports $\mathrm{ARS}_{\text{net}}$ (\textsc{flip})
alongside $\Delta_{\text{net}}$ over all $500$ machine documents
(\textsc{unc}), together with machine-class recall (\textsc{rec}).The conditioned $\mathrm{ARS}_{\text{net}}$ is consistently larger than $\Delta_{\text{net}}$,
as the selection argument of Section~\ref{method:counterfactual} predicts.  For \texttt{chatgpt-detector-roberta}, re-wrapping the lines of a machine-written arXiv abstract reverses $42$-$58\%$ of its correct decisions relative to the placebo; the unconditioned
margin is $+0.20$ to $+0.27$. The three detectors rank as following: the HC3 detector relies most on formatting, OpenAI's GPT-2 detector less, and the
adversarially trained RADAR does not rely at all ($-0.004$ on the arXiv subsets).

\paragraph{Is this a selection artifact?}
IDMGSP provides a control that no confound can imitate. It is the one subset where the artifact is reversed: machine introductions carry newlines and human ones do not
(Section~\ref{diag:whitespace}). If the measurement reflected selection, it could only bias
towards positive values, because the documents a weak detector classifies correctly are the ones
already most machine-like. Instead, \texttt{chatgpt-detector-roberta} reverses sign on IDMGSP,
significantly and without conditioning ($-0.171$, CI $[-0.206, -0.138]$). Since IDMGSP's newlines appear mostly in machine texts, it is the machine profile that carries
the human-looking cue, and imposing it flips $41.8\%$ of correct decisions against
$7.9\%$ for the human profile. RADAR, which does not rely on formatting, shows no comparable reversal. Together with
the null on CHEAT, whose classes do not differ in formatting at all ($-0.010$, CI spanning zero),
this is the signature of directional sensitivity to the formatting profile.

The residual dependence on the size of the conditioning set is not
significant for \texttt{chatgpt-detector-roberta} ($\rho = -0.430$, $p = 0.21$ between $|T_M|$ and
the unconditioned margin). Though the effect is one-directional: imposing the machine profile on correctly classified human documents flips at most 2\% of decisions for any detector. All three treat wrapped text as evidence of human authorship, but none treats unwrapped text as evidence of machine authorship. The same one-sidedness shows in detectors' recall, which is above $0.95$ on human documents and between $0.20$ and $0.98$ on machine ones.

\subsection{Meaning of Diagnostic}
\label{res:validation}

The probe predicts artifact reliance. Across the ten subsets, each diagnostic's per-subset value correlates (Spearman, $n=10$) with \texttt{chatgpt-detector-roberta}'s $\mathrm{ARS}_{\text{net}}$: the full probe's macro-$F_1$ gives $\rho = 0.939$ ($p = 0.0001$); the formatting-only AUC gives $\rho = 0.770$ ($p = 0.0092$); and the normalization drop gives $\rho = 0.673$ ($p = 0.0330$). The lexical TF-IDF reference does not ($\rho = -0.552$, $p = 0.098$), so the relationship is specific to surface form rather than to separability in general. A cheap, content-free probe run at release time therefore flags the subsets on which a deployed detector will later prove format-dependent, which is a useful check on its own.

It does not predict cross-benchmark degradation. Against the decontaminated
generalization gap $\Delta_{\text{gen}}$ of Section~\ref{res:transfer}, every diagnostic we
tested is uninformative and, if anything, pointed the wrong way: probe macro-$F_1$
$\rho = -0.164$ ($p = 0.65$), formatting-only AUC $\rho = -0.236$ ($p = 0.51$),
$\mathrm{ARS}_{\text{net}}$ $\rho = -0.333$ ($p = 0.35$). We report this as a negative result.
However, surface separability and transfer failure are distinct properties of a benchmark, and the fact that a
benchmark is easy to game does not by itself predict that detectors trained on it will generalize poorly.

With $n = 10$ these correlations support a qualitative conclusion rather than a precise
estimate; extending the unit of analysis to (subset $\times$ generator) cells would give more
power and is left to future work.

\subsection{Transfer and Normalization}
\label{res:transfer}

Both the probe and the public detectors were assessed within a single benchmark. This section asks whether relying on it can help to perform out-of-domain. We fine-tuned RoBERTa and DeBERTa on each subset and tested on the other nine. After that we asked three questions of the same runs: does the artifact survive in transfer to another benchmark; does intra-benchmark duplication inflate the scores we report; and does removing the artifact from training data, via normalization or reweighting, close the exploit?

Before running the experiments, we excluded
test documents that share a near-duplicate with the training subset. We ran two backbones, because the artifact turned out to
be visible to one and invisible to the other.

\paragraph{Exploitability depends on the tokenizer.}
\texttt{DeBERTa-v3-base} uses SentencePiece, which collapses whitespace runs and discards
newlines: raw and normalized documents yield identical token sequences, so the model
cannot represent the cue that dominates these benchmarks. \texttt{roberta-base} uses a byte-pair encoding that preserves both. In-domain macro-$F_1$ under the cluster-aware split is showed in Table~\ref{tab:tokenizer}.

\begin{table}[h!]
\centering
\small
\setlength{\tabcolsep}{4pt} 
\renewcommand{\arraystretch}{0.95}
\caption{In-domain macro-$F_1$ on the arXiv subsets under the cluster-aware
split: the content-free surface probe, RoBERTa (byte-pair encoding, sees the
whitespace cue) and DeBERTa-v3 (SentencePiece, discards it).}
\label{tab:tokenizer}
\begin{tabular}{lccc}
\toprule
Subset & surface probe & RoBERTa & DeBERTa \\
\midrule
M4 arXiv        & 0.995 & \textbf{0.996} & 0.687 \\
SemEval24 arXiv & 0.997 & \textbf{0.997} & 0.711 \\
COLING25 arXiv  & 0.951 & \textbf{0.995} & 0.773 \\
\bottomrule
\end{tabular}
\end{table}

The byte-pair model behaves similarly with the content-free probe; the SentencePiece model performs instead at
the level of a probe with its formatting features removed ($0.729$-$0.752$,
Table~\ref{tab:probe}, column Norm.). A benchmark artifact is therefore
not equally available to all detecting systems, and scores on these resources are not comparable across tokenizer families: a benchmark can be trivial for one architecture and appear sophisticated to another. 

On the raw text, the byte-pair model reaches $0.995$-$0.997$ in-domain on the arXiv subsets and $0.333$ on every one of the four format-clean benchmarks. $0.333$ is the macro-$F_1$ of predicting a single class on a balanced test
set, i.e.\ it is a complete collapse of the model, that indicates that detector has simple learned to count line breaks and cannot be generalized to the texts without this exploit.

\paragraph{Normalization trades in-domain score for generalization.}
If we normalize the text before training, macro-$F_1$ in-domain in aggregate drops by $0.074$ ($0.225$ on the arXiv subsets), meanwhile decontaminated out-of-domain macro-$F_1$ increases by $0.030$, reducing the generalization gap by a third, from $0.320$ to $0.216$ (Table~\ref{tab:transfer}).
When we restrict to the transfers that are out-of-domain such as arXiv-trained models evaluated on
the four disjoint benchmarks,  the effect is much larger: mean macro-$F_1$ rises from $0.333$ to
$0.652$, a gain of $+0.319$, with per-subset gains of $+0.411$ (COLING25 arXiv), $+0.347$
(SemEval24 arXiv) and $+0.197$ (M4 arXiv). Subsets without a formatting artifact barely change (CHEAT from $0.958$ to $0.964$, GPABench2 CS from $0.995$ to $0.995$, IDMGSP
from $0.965$ to $0.958$), so the intervention is not a general regulariser.

Replacing the random split with a cluster-aware one costs $0.005$ macro-$F_1$ for RoBERTa and
$0.006$ for DeBERTa. We noticed significant duplication within some of the subsets, however, as it did not measurably inflate reported scores on either backbone, we simply note it, see more details in Appendix~\ref{appendix:duplication}.

\paragraph{Artifact-debiased training.}
We tried to run artifact-debiased training by reweighting examples by the surface bias model and report the results in Table~\ref{tab:transfer}. It degraded both in-domain
($0.981$ to $0.836$) and out-of-domain ($0.662$ to $0.657$), and is
worse than normalization alone on both axes. On the arXiv subsets the bias model reaches $p_{\text{bias}} \approx 0.99$ using
whitespace, so almost every example receives near-zero weight, collapsing to majority-class prediction ($0.333 \pm 0.000$ for both backbones) as on SemEval24 arXiv. Where the bias model is informative but not saturated, reweighting does improve out-of-domain performance: GPABench2 CS $0.733 \rightarrow 0.748$, Physics $0.748 \rightarrow 0.751$, IDMGSP $0.669 \rightarrow 0.684$.
We conclude that bias-model reweighting requires a bias model that is neither saturated nor operating on features outside the detector's representational reach, because for example DeBERTa tokenizer discards part of the features, so their effect here is near-null.

\begin{table}[t]
\centering\small
\caption{Performance of fine-tuned detectors. 
$\Delta_{\text{gen}}$ is in-domain minus out-of-domain. \textsc{cluster} keeps near-duplicate groups on one side of the split; \textsc{norm.}
additionally normalizes surface form; \textsc{debias} adds bias-model reweighting.}
\label{tab:transfer}
\centering
\small 
\setlength{\tabcolsep}{4pt} 
\renewcommand{\arraystretch}{0.95}
\begin{tabular}{llccc}
\toprule
Backbone & Cond. & in-domain & OOD & $\Delta_{\text{gen}}$ \\
\midrule
\multirow{4}{*}{RoBERTa}
 & \textsc{rand}, raw   & 0.986{\tiny$\pm$0.004} & 0.666{\tiny$\pm$0.004} & 0.320 \\
 & \textsc{clust}, raw  & 0.981{\tiny$\pm$0.003} & 0.662{\tiny$\pm$0.014} & 0.320 \\
 & \ +\textsc{norm.} & 0.907{\tiny$\pm$0.009} & \textbf{0.692}{\tiny$\pm$0.010} & \textbf{0.216} \\
 & \ +\textsc{debias} & 0.836{\tiny$\pm$0.010} & 0.657{\tiny$\pm$0.018} & 0.179 \\
\midrule
\multirow{4}{*}{DeBERTa}
 & \textsc{rand}, raw   & 0.884{\tiny$\pm$0.030} & 0.653{\tiny$\pm$0.024} & 0.231 \\
 & \textsc{clust}, raw  & 0.878{\tiny$\pm$0.021} & 0.649{\tiny$\pm$0.012} & 0.229 \\
 & \ +\textsc{norm.} & 0.878{\tiny$\pm$0.026} & 0.651{\tiny$\pm$0.022} & 0.227 \\
 & \ +\textsc{debias} & 0.773{\tiny$\pm$0.035} & 0.597{\tiny$\pm$0.019} & 0.175 \\
\bottomrule
\end{tabular}
\end{table}

\section{Discussion}
\label{sec:discussion}

On the three arXiv subsets, a logistic
regression over content-free surface features outperforms a fine-tuned transformer by
$0.18$-$0.31$ macro-$F_1$: whatever these benchmarks measure, it is not the property they
were built to measure, since they can be played with a solution that requires no access to the lexical content.
And since this cue is visible only to some tokenizer families
(Section~\ref{res:transfer}), some systems will have an advantage unrelated to how good the system actually is, but rather to whether or not it inadvertently erases exploitable artifacts. We are, however, not aware of any benchmark in this area that
documents which family it advantages.

The defect comes from document processing, which makes it cheap to prevent. Human documents should be normalized to the same surface form as generations before release. We recommend shipping the surface-probe score and per-feature AUCs  with any such resource. 


\section{Conclusion}

We audited ten scientific-document subsets of six machine-generated-text detection benchmarks and found that much of their reported difficulty is an artifact of document assembly: a logistic regression over content-free surface features matches or exceeds the corresponding shared-task winners, and on the arXiv subsets the dominant cue is whitespace inherited from how human documents were collected. Format counterfactuals and the placebo-corrected Artifact Reliance Score separate benchmarks that contain such a cue from detectors that rely on one, and whether the cue is exploitable at all depends on the detector's tokenizer. We recommend normalizing surface form before release and publishing a surface-probe score alongside every resource.

\section*{Limitations}

We analyzed only ten subsets of six resources, all English and all of scientific domain. We do not claim the measured effects will transfer to other domains or languages, and the formatting artifact in particular is a property of specific collection pipelines rather than of machine-generated text, although the reason it appears comes from a widely used processing pipeline, so we believe it should be quite common.

$\mathrm{ARS}_{\text{net}}$ is reported for three publicly available detectors. Because it
conditions on correct decisions, its denominator shrinks with a detector's recall, and the detectors we measured have machine-class recall between $0.20$ and $0.978$ on this data; the score should therefore not be used to rank detectors of very unequal accuracy. Our correlation analyzes use $n = 10$ subsets, which supports qualitative conclusions but not precise estimates.

The fine-tuned baseline is a single architecture at one size, trained for three epochs with
best-epoch selection over three seeds. Its in-domain scores on the arXiv subsets
($0.769$-$0.824$) are well below the corresponding shared-task winners, and although we
attribute this to the tokenizer's inability to represent the dominant cue, we cannot exclude
that a larger or better-tuned model would score higher. Finally, our
counterfactual operators cover whitespace and punctuation style only, and other structural operators are considered as future work. 

\bibliography{custom}

\appendix

\section{Implementation and Hyperparameters}
\label{appendix:hyperparameters}

All randomness is seeded. A single split seed ($0$) is shared by every condition and model so
that runs differ only in the factor under study; training seeds are $\{0,1,2\}$. Splits are formed in two ways:
randomly, as is standard, and cluster-aware, so that no near-duplicate group spans the
train/test boundary. Transfer evaluations exclude test documents having a near-duplicate in the
training subset.

\paragraph{Surface probe (Section~\ref{method:probe}).}
$27$ features per document: four size features (characters, words, sentences, mean sentence
length), four word-shape features (mean and standard deviation of word length, type--token
ratio, hapax rate), ten per-character punctuation rates, four formatting features
(\texttt{r\_newline}, \texttt{r\_ws\_run}, \texttt{r\_curly\_quote}, \texttt{r\_fancy\_dash}),
and five further surface markers (non-ASCII, digit and uppercase rates, numeric-token and
LaTeX-command rates). For classifier, we put these standardized features into logistic regression
($\texttt{max\_iter}=5000$, L2, default $C$), evaluated by $5$-fold stratified
cross-validation on balanced samples of up to $10{,}000$ documents per class per subset.
Confidence intervals are $1000$-sample bootstrap percentiles over the out-of-fold predictions.
The lexical reference is TF-IDF over word $1$-$2$ grams, $50{,}000$ maximum features,
$\texttt{min\_df}=2$, sublinear term frequency, into the same classifier and folds. TF-IDF is
computed on raw text only: its default token pattern discards whitespace and standalone
punctuation, so normalization cannot alter any feature value, and we verified the two runs
are numerically identical.


\paragraph{Format counterfactuals and ARS (Section~\ref{method:counterfactual}).}
$500$ documents per class per subset, each in four variants (original, normalized,
human-profile injection, machine-profile placebo), giving $40{,}000$ scored documents per
detector. The admissibility invariant is asserted on every
transformed document. Formatting profiles are estimated per class per subset from line-length
distributions, double-space rate at sentence boundaries, and curly-quote and dash rates.
Detectors are scored with $\texttt{max\_length}=512$ and batch size $16$. 


\paragraph{Fine-tuned detector (Section~\ref{res:transfer}).}
\texttt{DeBERTa-v3-base}, 3 epochs, AdamW at learning rate $2\times10^{-5}$, linear decay with $10\%$
warmup, gradient-norm clipping at $1.0$, bfloat16 autocast on CUDA. Maximum sequence length is
$512$ for M4 peer reviews and IDMGSP introductions and $256$ for the remaining subsets, whose
medians are below $210$ tokens. Test sets are $1000$ balanced documents per subset, held fixed
across every condition and seed. Clustering for the cluster-aware split considers up to
$25{,}000$ documents per subset. The debiasing condition weights each training example by
$(1-p_{\text{bias}})^{\gamma}$ with $\gamma=1$, mean-normalized and then clipped at ten times the
mean. The bias model is the surface probe of Section~\ref{method:probe}, fitted on the
training split only. RoBERTa-base uses identical hyperparameters

\section{Evaluation results of competitions}
\label{appendix:evaluation}

Table~\ref{table:competition_results} lists the best reported score in each shared task whose data appears in Table~\ref{tab:subsets}. These are the figures we use to compare with surface probe (Section~\ref{method:probe}).

\begin{table}[h!]
\centering
\caption{Best reported score in each shared task represented in our analysis.}
\label{table:competition_results}
\begin{tabular}{lcc}
\toprule
\textbf{Competition} & \textbf{Metric} & \textbf{Best result} \\ 
\midrule
SemEval24 Mono                     & Accuracy                                     & 0.975                                \\ \hline
SemEval24  Multi & Accuracy                                     & 0.959                                \\ \hline
COLING25 Mono & Macro-F$_1$ & 0.8307 \\
\hline
COLING25 Multi & Macro-F$_1$   & 0.7916 \\
\bottomrule
\end{tabular}

\end{table}

\section{Intra-Benchmark Duplication}
\label{appendix:duplication}

Table~\ref{tab:intradup} reports duplication within each subset. We observe that M4 arXiv contains $2{,}999$ duplicate groups of about five members each, summing up to $15{,}002$ rows, which is precisely its $15{,}000$ human documents, and each human abstract is stored once per generator. 

The natural expectation is that this inflates reported scores, since a random split places copies of the same document on both sides of it. We tested this directly and find no measurable effect. Replacing the random split with a cluster-aware split that keeps every near-duplicate group completely within one side changes in-domain macro-$F_1$ by 0.005/ 0.006. the sign is inconsistent across subsets, and a subset's
duplicate rate does not predict its change ($\rho = 0.34$, $p = 0.34$). Even M4 arXiv, at $35\%$
duplication, moves only from $0.740$ to $0.687$, within seed variance. We therefore describe duplication as a defect in effective dataset size and in the interpretation of a nominally balanced split, and not as a source of score inflation.

\begin{table}[h]
\centering
\caption{Duplication within each subset. Near-duplicates use MinHash with Jaccard $\ge 0.8$;
exact matching is on whitespace-normalized text.}
\label{tab:intradup}
\begin{tabular}{lrr}
\toprule
\textbf{Subset} & \textbf{exact} & \textbf{near} \\
\midrule
M4 PeerRead        & 40.0\% & 50.0\% \\
M4 arXiv           & 35.2\% & 49.5\% \\
COLING25 PeerRead  & 28.0\% & 55.1\% \\
COLING25 arXiv     &  7.7\% & 15.7\% \\
SemEval24 PeerRead &  1.3\% &  2.7\% \\
SemEval24 arXiv    &  0.00\% &  0.01\% \\
IDMGSP             &  0.03\% &  0.24\% \\
CHEAT, GPABench2   & $\le$0.01\% & $\le$0.03\% \\
\bottomrule
\end{tabular}
\end{table}

\section{Subset Construction Details}
\label{appendix:data}



Table~\ref{tab:subsets} reports abbreviated generator lists. For the M4 family, scientific subsets were isolated by matching the release's domain field against arXiv and PeerRead. Documents shorter than five whitespace tokens were dropped as degenerate. No class rebalancing was applied at isolation time; the balanced samples used in each experiment are drawn afterwards with a fixed seed.
\begin{table*}[h!]
\centering
\small
\caption{The ten scientific-document subsets analyzed in this work. $|H|$ and $|M|$ are
counts of human and machine documents in the isolated subset; median length is in
whitespace tokens. All subsets are English and all machine documents are written from
scratch. Generator counts are as recorded in each release.}
\label{tab:subsets}
\resizebox{\linewidth}{!}{%
\begin{tabular}{llrrccl}
\toprule
\textbf{Subset} & \textbf{Document type} & $\mathbf{|H|}$ & $\mathbf{|M|}$ &
\textbf{\makecell[c]{Median tokens\\ H \textbf{/} M}} & \textbf{\#Gen.} &
\textbf{Generators} \\
\midrule
\multicolumn{7}{l}{\textit{M4-based datasets}} \\
\midrule
M4 arXiv & abstracts & 15{,}000 & 15{,}000 & 198 \textbf{/} 180 & 5 &
  \makecell[l]{gpt-3.5-turbo, text-davinci-003,\\ cohere, dolly-v2-12b, flan-t5} \\
\hline
M4 PeerRead & peer reviews & 2{,}930 & 2{,}930 & 2173 \textbf{/} 1026 & 5 &
  \makecell[l]{chatgpt, davinci, cohere,\\ dolly-v2, flan-t5} \\
\hline
SemEval24 arXiv & abstracts & 15{,}498 & 11{,}999 & 194 \textbf{/} 194 & 4 &
  chatGPT, davinci, cohere, dolly \\
\hline
SemEval24 PeerRead & peer reviews & 2{,}357 & 9{,}374 & 197 \textbf{/} 238 & 4 &
  chatGPT, davinci, cohere, dolly \\
\hline
COLING25 arXiv & abstracts & 22{,}484 & 30{,}684 & 192 \textbf{/} 178 & 12 &
  \makecell[l]{+ gpt4o, gpt4, llama3-70b/8b,\\ gemma2-9b, gemma-7b, mixtral-8x7b,\\ bloomz} \\
\hline
COLING25 PeerRead & peer reviews & 3{,}300 & 16{,}169 & 203 \textbf{/} 207 & 7 &
  \makecell[l]{+ gpt4o, gpt4, bloomz} \\
\midrule
\multicolumn{7}{l}{\textit{Research-paper resources}} \\
\midrule
IDMGSP & paper introductions & 8{,}000 & 8{,}000 & 423 \textbf{/} 237 & 5 &
  \makecell[l]{SCIgen, GPT-2, GPT-3,\\ ChatGPT, Galactica} \\
\hline
GPABench2 CS & abstracts & 50{,}000 & 200{,}000 & 153 \textbf{/} 164 & 1 &
  ChatGPT ($4$ prompt templates) \\
\hline
GPABench2 Physics & abstracts & 50{,}000 & 200{,}000 & 148 \textbf{/} 155 & 1 &
  ChatGPT ($4$ prompt templates) \\
\hline
CHEAT & abstracts (IEEE) & 15{,}395 & 15{,}395 & 156 \textbf{/} 131 & 1 &
  ChatGPT \\
\bottomrule
\end{tabular}%
}
\end{table*}

\end{document}